\documentclass[10pt,twocolumn,letterpaper]{article}

\usepackage{iccv}
\usepackage{times}
\usepackage{epsfig}
\usepackage{graphicx}
\usepackage{amsmath}
\usepackage{amssymb}

\usepackage[pagebackref=true,breaklinks=true,letterpaper=true,colorlinks,bookmarks=false]{hyperref}

\iccvfinalcopy % *** Uncomment this line for the final submission

\def\iccvPaperID{1101} % *** Enter the ICCV Paper ID here

\ificcvfinal\fi
\begin{document}

%%%%%%%%% TITLE
\title{Unmasking the abnormal events in video\vspace*{-0.1cm}}

\author{Radu Tudor Ionescu$^{1,2}$, Sorina Smeureanu$^{1,2}$, Bogdan Alexe$^{1,2}$, Marius Popescu$^{1,2}$\\
$^1$University of Bucharest, 14 Academiei, Bucharest, Romania\\
%{\tt\small raducu.ionescu@gmail.com},
%{\tt\small bogdan.alexe@fmi.unibuc.ro},
%{\tt\small popescunmarius@gmail.com}
$^2$SecurifAI, 24 Mircea Vod\u{a}, Bucharest, Romania\vspace*{-0.2cm}
}

\maketitle
% \thispagestyle{empty}

%%%%%%%%% ABSTRACT
\begin{abstract}
We propose a novel framework for abnormal event detection in video that requires no training sequences. Our framework is based on unmasking, a technique previously used for authorship verification in text documents, which we adapt to our task. We iteratively train a binary classifier to distinguish between two consecutive video sequences while removing at each step the most discriminant features. Higher training accuracy rates of the intermediately obtained classifiers represent abnormal events. To the best of our knowledge, this is the first work to apply unmasking for a computer vision task.  We compare our method with several state-of-the-art supervised and unsupervised methods on four benchmark data sets. The empirical results indicate that our abnormal event detection framework can achieve state-of-the-art results, while running in real-time at $20$ frames per second.
\end{abstract}

\vspace*{-0.4cm}
%%%%%%%%% BODY TEXT
\section{Introduction}
\begin{figure}[t]

\begin{center}
\includegraphics[width=1.0\columnwidth]{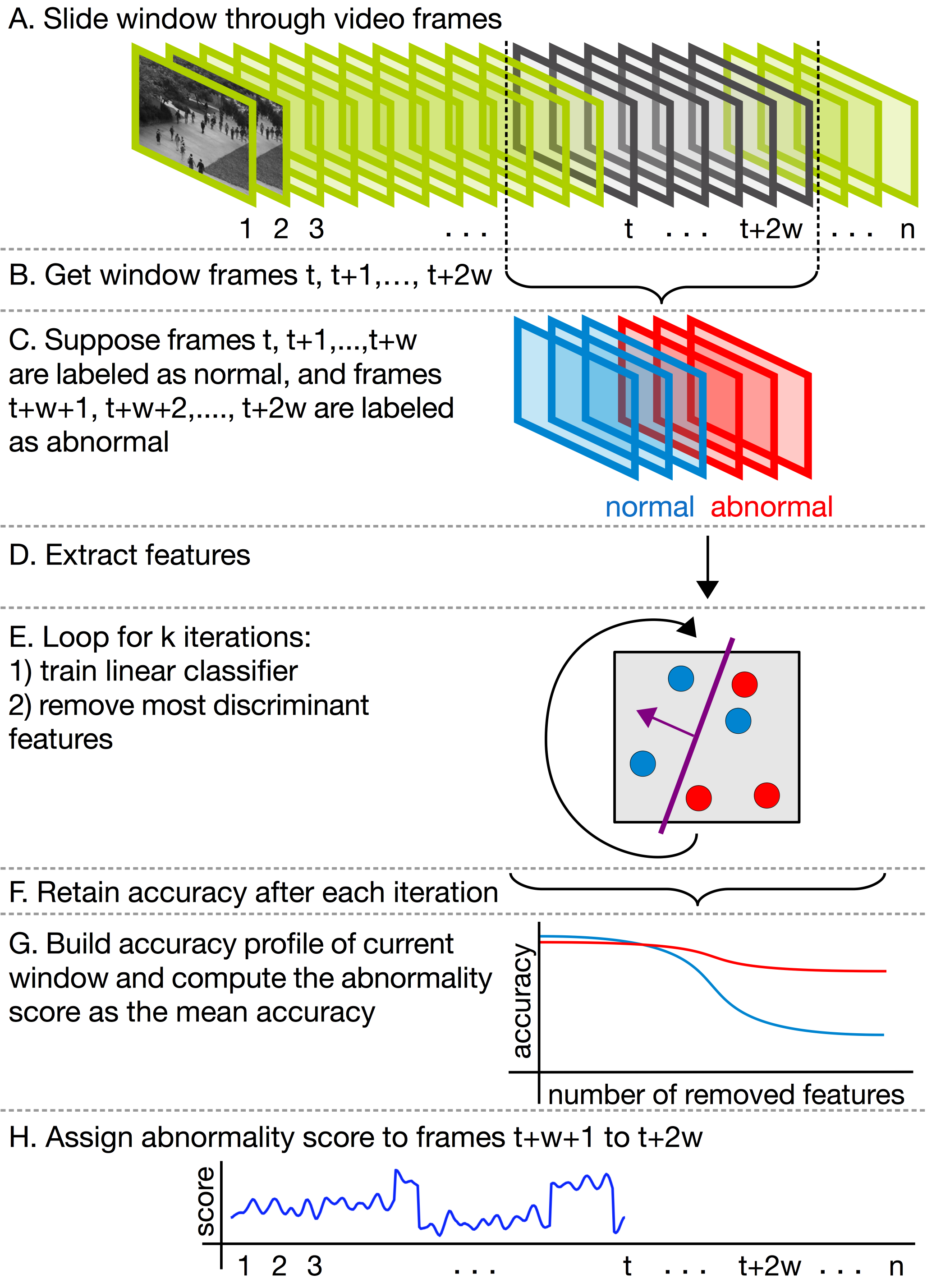}
\end{center}
\vspace*{-0.4cm}
\caption{Our anomaly detection framework based on unmasking~\cite{Koppel-JMLR-2007}. The steps are processed in sequential order from (A) to (H). Best viewed in color.}
\label{fig_pipeline}
\vspace*{-0.4cm}
\end{figure}

Abnormal event detection in video is a challenging task in computer vision, as the definition of what an abnormal event looks like depends very much on the context. For instance, a car driving by on the street is regarded as a normal event, but if the car enters a pedestrian area, this is regarded as an abnormal event. A person running on a sports court (normal event) versus running outside from a bank (abnormal event) is another example. Although what is considered abnormal depends on the context, we can generally agree that abnormal events should be unexpected events~\cite{Itti-CVPR-2005} that occur less often than familiar (normal) events. As it is generally impossible to find a sufficiently representative set of anomalies, the use of traditional supervised learning methods is usually ruled out. Hence, most abnormal event detection approaches~\cite{Antic-ICCV-2011,Cheng-CVPR-2015,Kim-CVPR-2009,Li-PAMI-2014,Lu-ICCV-2013,Mahadevan-CVPR-2010,Mehran-CVPR-2009,Xu-BMVC-2015,Zhao-CVPR-2011} learn a model of familiarity from a given training video and label events as abnormal if they deviate from the model. In this paper, we consider an even more challenging setting, in which no additional training sequences are available~\cite{Giorno-ECCV-2016}. As in this setting we cannot build a model in advance and find deviations from it, our approach is completely unsupervised, as we briefly explain next. Our method labels a short-lasting event as abnormal if the amount of change from the immediately preceding event is substantially large. We quantify the change as the training accuracy of a linear classifier applied on a sliding window that comprises both the preceding and the currently examined event, as illustrated in Figure~\ref{fig_pipeline}. We consider that the first half of the window frames are labeled as normal and take them as reference. We suppose the second half are labeled as abnormal, but we seek to find if this hypothesis is indeed true. We extract both motion and appearance features from the frames, and train a binary classifier with high regularization to distinguish between the labeled frames. We retain the training accuracy of the classifier and repeat the training process by eliminating some of the best features. This process is known as \emph{unmasking} \cite{Koppel-JMLR-2007} and it was previously used for authorship verification of text documents. To the best of our knowledge, we are the first to apply unmasking for a computer vision task. After a certain number of iterations with unmasking, we can build a profile (plot) with the collected accuracy rates in order to assess if the current event, represented by the second half of the frames, does contain enough changes to consider it abnormal. Intuitively, if the change is significant, the classification accuracy should stay high even after eliminating a certain amount of discriminating features. Otherwise, the accuracy should drop much faster as the discriminating features get eliminated, since the classifier will have a hard time separating two consecutive normal events. We estimate the accuracy profile obtained by unmasking with the mean of the accuracy rates, and consider the mean value to represent the anomaly score of the frames belonging to the current event.

We perform abnormal event detection experiments on the Avenue \cite{Lu-ICCV-2013}, the Subway \cite{Adam-PAMI-2008}, the UCSD \cite{Mahadevan-CVPR-2010} and the UMN \cite{Mehran-CVPR-2009} data sets in order to compare our unsupervised approach with a state-of-the-art unsupervised method \cite{Giorno-ECCV-2016} as well as several supervised methods \cite{Cong-CVPR-2011,Kim-CVPR-2009,Lu-ICCV-2013,Mahadevan-CVPR-2010,Mehran-CVPR-2009,Ren-BMVC-2015,Saligrama-CVPR-2012,Sun-PR-2017,Xu-BMVC-2015,Zhang-PR-2016}. The empirical results indicate that we obtain better results than the unsupervised approach \cite{Giorno-ECCV-2016} and, on individual data sets, we reach or even surpass the accuracy levels of some supervised methods \cite{Cong-CVPR-2011,Kim-CVPR-2009,Lu-ICCV-2013,Mehran-CVPR-2009}. Unlike the approach of \cite{Giorno-ECCV-2016}, our method can process the video in real-time at $20$ frames per second.

We organize the paper as follows. We present related work on abnormal event detection in Section~\ref{sec_RelatedWork}. We describe our unsupervised learning framework in Section~\ref{sec_Method}. We present the abnormal event detection experiments in Section~\ref{sec_Experiments}. Finally, we draw our conclusions in Section~\ref{sec_Conclusion}.

\section{Related Work}
\label{sec_RelatedWork}

Abnormal event detection is usually formalized as an outlier detection task~\cite{Antic-ICCV-2011,Cheng-CVPR-2015,Cong-CVPR-2011,Dutta-AAAI-2015,Kim-CVPR-2009,Li-PAMI-2014,Lu-ICCV-2013,Mahadevan-CVPR-2010,Mehran-CVPR-2009,Ren-BMVC-2015,Sun-PR-2017,Xu-BMVC-2015,Zhang-PR-2016,Zhao-CVPR-2011}, in which the general approach is to learn a model of normality from training data and consider the detected outliers as abnormal events. Some abnormal event detection approaches~\cite{Cheng-CVPR-2015,Cong-CVPR-2011,Dutta-AAAI-2015,Lu-ICCV-2013,Ren-BMVC-2015} are based on learning a dictionary of normal events, and label the events not represented by the dictionary as abnormal. Other approaches have employed deep features~\cite{Xu-BMVC-2015} or locality sensitive hashing filters~\cite{Zhang-PR-2016} to achieve better results.

% \subsection{Key Differences from Similar Approaches}
There have been some approaches that employ unsupervised steps for abnormal event detection~\cite{Dutta-AAAI-2015,Ren-BMVC-2015,Sun-PR-2017,Xu-BMVC-2015}, but these approaches are not fully unsupervised. The approach presented in~\cite{Dutta-AAAI-2015} is to build a model of familiar events from training data and incrementally update the model in an unsupervised manner as new patterns are observed in the test data. In a similar fashion, Sun et al.~\cite{Sun-PR-2017} train a Growing Neural Gas model starting from training videos and continue the training process as they analyze the test videos for anomaly detection. Ren et al.~\cite{Ren-BMVC-2015} use an unsupervised approach, spectral clustering, to build a dictionary of atoms, each representing one type of normal behavior. Their approach requires training videos of normal events to construct the dictionary. Xu et al.~\cite{Xu-BMVC-2015} use Stacked Denoising Auto-Encoders to learn deep feature representations in a unsupervised way. However, they still employ multiple one-class SVM models to predict the anomaly scores. 

To the best of our knowledge, the only work that does not require any kind of training data for abnormal event detection is~\cite{Giorno-ECCV-2016}. The approach proposed in~\cite{Giorno-ECCV-2016} is to detect changes on a sequence of data from the video to see which frames are distinguishable from all the previous frames. As the authors want to build an approach independent of temporal ordering, they create shuffles of the data by permuting the frames before running each instance of the change detection. Our framework is most closely related to~\cite{Giorno-ECCV-2016}, but there are several key differences that put a significant gap between the two approaches. % which we detail next. 
An important difference is that our framework is designed to process the video online, as expected for practical real-world applications. Since the approach of Del Giorno et al.~\cite{Giorno-ECCV-2016} needs to permute the test video frames before making a decision, the test video can only be processed offline. As they discriminate between the frames in a short window and all the frames that precede the window, their classifier will require increasingly longer training times as the considered window reaches the end of the test video. In our case, the linear classifier requires about the same training time in every location of the video, as it only needs to discriminate between the first half of the frames and the second half of the frames within the current window. Moreover, we train our classifier in several loops by employing the unmasking technique. Del Giorno et al.~\cite{Giorno-ECCV-2016} use the same motion features as~\cite{Lu-ICCV-2013}. We also use spatio-temporal cubes~\cite{Lu-ICCV-2013} to represent motion, but we remove the Principal Component Analysis (PCA) step for two reasons. First of all, we need as many features as we can get for the unmasking technique which requires more features to begin with. Second of all, training data is required to learn the PCA projection. Different from~\cite{Giorno-ECCV-2016}, we additionally use appearance features from pre-trained convolutional neural networks~\cite{Chatfield-BMVC-14}. With all these distinct characteristics, our framework is able to obtain better performance in terms of accuracy and time, as shown in Section~\ref{sec_Experiments}.

\section{Method}
\label{sec_Method}

We propose an abnormal event detection framework based on unmasking, that requires no training data. Our anomaly detection framework is comprised of eight major steps, which are indexed from A to H in Figure~\ref{fig_pipeline}. We next provide an overview of our approach, leaving the additional details about the non-trivial steps for later. We first apply a sliding window algorithm (step A) and, for each window of $2\cdot w$ frames (step B), we suppose that the first $w$ frames are normal and the last $w$ frames are abnormal (step C). After extracting motion or appearance features (step D), we apply unmasking (steps E to G) by training a classifier and removing the highly weighted features for a number of $k$ loops. We take the accuracy rates after each loop (step F) and build the accuracy profile of the current window (step G). Abnormal events correspond to high (almost constant) accuracy profiles (depicted in red), while normal events correspond to dropping accuracy profiles (depicted in blue). We compute the anomaly score for the last $w$ frames as the mean of the retained accuracy rates (step H).

For the sake of simplicity, there are several important aspects that are purposely left out in Figure~\ref{fig_pipeline}. 
First of all, we divide the frames into $2 \times 2$ spatial bins, thus obtaining four sub-videos, which we process individually through our detection framework until step G. Hence, for each video frame, we produce four anomaly scores, having one score per bin. Before step H, we assign the score of each frame as the maximum of the four anomaly scores corresponding to the $2 \times 2$ bins. 
Second of all, we apply the framework independently using motion features on one hand and appearance features on the other. For each kind of features, we divide the video into $2 \times 2$ bins and obtain a single anomaly score per frame as detailed above. To combine the anomaly scores from motion and appearance features, we employ a late fusion strategy by averaging the scores for each frame, in step H.
Third of all, we take windows at a predefined interval $s$ (stride), where the choice of $s$ can generate overlapping windows (e.g. $s=1$ and $w=10$). In this situation, the score of a frame is obtained by averaging the anomaly scores obtained after processing every separate window that includes the respective frame in its second half. We apply a Gaussian filter to temporally smooth the final anomaly scores. % frame-level anomaly scores.
% An algorithm can be formally presented here.
We present additional details about the motion and appearance features (step D) in Section~\ref{sec_Features}, and about the unmasking approach (steps E to G) in Section~\ref{sec_Unmasking}.

% When we consider a stride that generates overlapping windows, we average the anomaly scores (obtained after processing each window) for each frame.

\subsection{Features}
\label{sec_Features}

Unlike other approaches~\cite{Cheng-CVPR-2015,Xu-BMVC-2015}, we apply the same steps in order to extract motion and appearance features from video, irrespective of the data set.

\noindent
{\bf Motion features.}
Given the input video, we resize all frames to $160 \times 120$ pixels and uniformly partition each frame to a set of non-overlapping $10 \times 10$ patches. Corresponding patches in $5$ consecutive frames are stacked together to form a spatio-temporal cube, each with resolution $10 \times 10 \times 5$. We then compute 3D gradient features on each spatio-temporal cube and normalize the resulted feature vectors using the $L_2$-norm. To represent motion, we essentially employ the same approach as~\cite{Giorno-ECCV-2016,Lu-ICCV-2013}, but without reducing the feature vector dimension from $500$ to $100$ components via PCA. This enables us to keep more features for unmasking. Since unmasking is about gradually eliminating the discriminant features, it requires more features to begin with. As~\cite{Giorno-ECCV-2016,Lu-ICCV-2013}, we eliminate the cubes that have no motion gradients (the video is static in the respective location). We divide the frames into $2 \times 2$ spatial bins of $80 \times 60$ pixels each, obtaining at most $48$ cubes per bin. Bins are individually processed through our detection framework. It is important to mention that each spatio-temporal cube is treated as an example in step E (Figure~\ref{fig_pipeline}) of our framework. Although we classify spatio-temporal cubes as~\cite{Giorno-ECCV-2016}, we assign the anomaly score to the frames, not the cubes.

\noindent
{\bf Appearance features.}
In many computer vision tasks, for instance predicting image difficulty~\cite{img-difficulty-CVPR-2016}, higher level features, such as the ones learned with convolutional neural networks (CNN)~\cite{Hinton-NIPS-2012} are the most effective. To build our appearance features, we consider a pre-trained CNN architecture able to process the frames as fast as possible, namely VGG-f~\cite{Chatfield-BMVC-14}. Considering that we want our detection framework to work in real-time on a standard desktop computer, not equipped with expensive GPU, the VGG-f~\cite{Chatfield-BMVC-14} is an excellent choice as it can process about $20$ frames per second on CPU. We hereby note that better anomaly detection performance can probably be achieved by employing deeper CNN architectures, such as VGG-verydeep~\cite{Simonyan-ICLR-14}, GoogLeNet~\cite{Szegedy-CVPR-2015} or ResNet~\cite{He-CVPR-2016}.

The VGG-f model is trained on the ILSVRC benchmark~\cite{Russakovsky2015}. It is important to note that fine-tuning the CNN for our task is not possible, as we are not allowed to use training data in our unsupervised setting. Hence, we simply use the pre-trained CNN to extract deep features as follows. Given the input video, we resize the frames to $224 \times 224$ pixels. We then subtract the mean imagine from each frame and provide it as input to the VGG-f model. We remove the fully-connected layers (identified as \emph{fc6}, \emph{fc7} and \emph{softmax}) and consider the activation maps of the last convolutional layer (\emph{conv5}) as appearance features. While the fully-connected layers are adapted for object recognition, the last convolutional layer contains valuable appearance and pose information which is more useful for our anomaly detection task. Ideally, we would like to have at least slightly different representations for a person walking versus a person running. From the \emph{conv5} layer, we obtain $256$ activation maps, each of $13 \times 13$ pixels. As for the motion features, we divide the activation maps into $2 \times 2$ spatial bins of $7 \times 7$ pixels each, such that the bins have a one-pixel overlap towards the center of the activation map. For each bin, we reshape the bins into $49$ dimensional vectors and concatenate the vectors corresponding to the $256$ filters of the \emph{conv5} layer into a single feature vector of $12544$ ($7 \times 7 \times 256$) components. The final feature vectors are normalized using the $L_2$-norm.

\subsection{Change Detection by Unmasking}
\label{sec_Unmasking}

The unmasking technique~\cite{Koppel-JMLR-2007} is based on testing the degradation rate of the cross-validation accuracy of learned models, as the best features are iteratively dropped from the learning process. Koppel et al.~\cite{Koppel-JMLR-2007} offer evidence that this unsupervised technique can solve the authorship verification problem with very high accuracy. We modify the original unmasking technique by considering the training accuracy instead of the cross-validation accuracy, in order to use this approach for \emph{online} abnormal event detection in video. We apply unmasking for each window of $2\cdot w$ frames at a stride $s$, where $w$ and $s$ are some input parameters of our framework. Our aim is to examine if the last $w$ frames in a given window represent an abnormal event or not. To achieve this purpose, we compare them with the first $w$ (reference) frames in the window. We assume that the first $w$ frames are labeled as normal and the last $w$ frames are labeled as abnormal, and train a linear classifier to distinguish between them. By training a classifier \emph{without unmasking}, we would only be able to determine if the first half of the window is distinguishable from the second half. Judging by the classifier's accuracy rate, we may consider that the last $w$ frames are abnormal if the accuracy is high and normal if the accuracy is low. This is essentially the underlying hypothesis of~\cite{Giorno-ECCV-2016}, with the difference that they assume all preceding frames (from the entire test video) as normal. Nevetheless, we consider only the immediately preceding $w$ frames as reference for our algorithm to run in real-time. As the number of normal (reference) samples is much lower, the classifier might often distinguish two normal events with high accuracy, which is not desired. Our \emph{main hypothesis} is that if two consecutive events are normal, then whatever differences there are between them will be reflected in only a relatively small number of features, despite possible differences in motion and appearance. Therefore, we need to apply unmasking in order to determine how large is the depth of difference between the two events. Similar to~\cite{Giorno-ECCV-2016}, we train a Logistic Regression classifier with high regularization. Different from~\cite{Giorno-ECCV-2016}, we eliminate the $m$ best features and repeat the training for a number of $k$ loops, where $m$ and $k$ are some input parameters of our framework. As the most discriminant features are gradually eliminated, it will be increasingly more difficult for the linear classifier to distinguish the reference examples, that belong to the first half of the window, from the examined examples, that belong to the second half. However, if the training accuracy rate over the $k$ loops drops suddenly, we consider that the last $w$ frames are normal according to our hypothesis. On the other hand, if the accuracy trend is to drop slowly, we consider that the analyzed frames are abnormal. Both kinds of accuracy profiles are shown in step G of Figure~\ref{fig_pipeline} for illustrative purposes, but, in practice, we actually obtain a single accuracy profile for a given window. In the end, we average the training accuracy rates over the $k$ loops and consider the average value to represent the degree of anomaly of the last $w$ frames in the window. We thus assign the same anomaly score to all the examined frames. It is interesting to note that Del Giorno et al.~\cite{Giorno-ECCV-2016} consider the probability that an example belongs to the abnormal class, hence assigning a different score to each example.

\vspace*{-0.2cm}
\section{Experiments}
\label{sec_Experiments}

\begin{table*}[t]
\small{
\begin{center}
\begin{tabular}{|l|c|c|c|c|c|r|r|}
\hline
Features 			& Bins 					& Unmasking 	& Stride     	& Frame AUC    	& Pixel AUC 	& Feature Extraction 	& \multicolumn{1}{|c|}{Prediction} \\
			 			& 		 					& 					 	& 			     	& 					     	& 					& \multicolumn{1}{|c|}{Time (FPS)}			& Time (FPS)\\
\hline
\hline
%VGG-f fc7			& $1\times 1$ 	& no 				& $1$			& $77.8\%$			& $92.5\%$ 			&	$21.4$		& $376.2$\\
%
%VGG-f fc6			& $1\times 1$ 	& no 				& $1$			& $77.9\%$			& $92.6\%$ 			&	$20.7$		& $376.3$\\
%
%VGG-f conv5		& $1\times 1$ 	& no 				& $1$			& $77.9\%$			& $92.7\%$ 			& $20.1$		& $59.8$\\
%
%VGG-f conv5		& $1\times 1$ 	& yes 				& $1$			& $79.2\%$			& $92.8\%$ 			&	$20.1$		& $9.8$\\
%
%VGG-f conv5		& $2\times 2$ 	& yes 				& $1$    	 	& $79.6\%$			& $92.9\%$ 			&	$20.1$		& $9.4$\\
%
%VGG-f conv5		& $2\times 2$ 	& yes 				& $2$			& $79.6\%$			& $93.0\%$ 			&	$20.1$		&	$18.3$\\
%
%VGG-f conv5		& $2\times 2$ 	& yes 				& $5$			& $79.6\%$			& $93.0\%$ 			&	$20.1$		& $42.2$\\
%
%VGG-f conv5		& $2\times 2$ 	& yes 				& $10$		& $79.5\%$			& $92.6\%$ 			&	$20.1$		& $78.1$\\
%\hline
%3D gradients	& $2\times 2$ 	& yes 				& $5$			& $78.7\%$			& $93.0\%$ 			&	$726.3$	& $34.9$\\
%
%VGG-f conv5 + 3D gradients	& $2\times 2$ & yes & $5$	& $79.7\%$			& $93.0\%$ 			&	$19.6$		& $19.3$\\

VGG-f fc7			& $1\times 1$ 	& no 				& $1$			& $78.3\%$			& $95.0\%$ 			&	$21.4$		& $376.2$\\

VGG-f fc6			& $1\times 1$ 	& no 				& $1$			& $78.4\%$			& $95.0\%$ 			&	$20.7$		& $376.3$\\

VGG-f conv5		& $1\times 1$ 	& no 				& $1$			& $78.6\%$			& $95.0\%$ 			& $20.1$		& $59.8$\\

VGG-f conv5		& $1\times 1$ 	& yes 				& $1$			& $81.1\%$			& $95.3\%$ 			&	$20.1$		& $9.8$\\

VGG-f conv5		& $2\times 2$ 	& yes 				& $1$    	 	& $82.5\%$			& $95.4\%$ 			&	$20.1$		& $9.4$\\

VGG-f conv5		& $2\times 2$ 	& yes 				& $2$			& $82.5\%$			& $95.4\%$ 			&	$20.1$		&	$18.3$\\

VGG-f conv5		& $2\times 2$ 	& yes 				& $5$			& $82.4\%$			& $95.4\%$ 			&	$20.1$		& $42.2$\\

VGG-f conv5		& $2\times 2$ 	& yes 				& $10$		& $82.0\%$			& $95.3\%$ 			&	$20.1$		& $78.1$\\
\hline
3D gradients	& $2\times 2$ 	& yes 				& $5$			& $79.8\%$			& $95.1\%$ 			&	$726.3$	& $34.9$\\

3D gradients + conv5 (late fusion)	& $2\times 2$ & yes & $5$	& $82.6\%$			& $95.4\%$ 			&	$19.6$		& $19.3$\\
\hline
\end{tabular}
\end{center}
}
\vspace*{-0.1cm}
\caption{Abnormal event detection results using various features, bins and window strides in our framework. The frame-level and the pixel-level AUC measures are computed on five test videos randomly chosen from the Avenue data set. For all models, the window size is $10$ and the regularization parameter is $0.1$. The number of frames per second (FPS) is computed by running the models on a computer with Intel Core i7 2.3 GHz processor and 8 GB of RAM using a single core.}
\label{tab_prelim_results}
\vspace*{-0.5cm}
\end{table*}

\vspace*{-0.1cm}
\subsection{Data Sets}

We show abnormal event detection results on four data sets. It is important to note that we use only the test videos from each data set, and perform anomaly detection \emph{without} using the training videos to build a model of normality.

\noindent
{\bf Avenue.}
We first consider the Avenue data set~\cite{Lu-ICCV-2013}, which contains $16$ training and $21$ test videos. In total, there are $15328$ frames in the training set and $15324$ frames in the test set. Each frame is $640 \times 360$ pixels. Locations of anomalies are annotated in ground truth pixel-level masks for each frame in the testing videos.

\noindent
{\bf Subway.}
One of the largest data sets for anomaly detection in video is the Subway surveillance data set~\cite{Adam-PAMI-2008}. It contains two videos, one of $96$ minutes (Entrance gate) and another one of $43$ minutes (Exit gate). The Entrance gate video contains $144251$ frames and the Exit gate video contains $64903$ frames, each with $512 \times 384$ resolution. Abnormal events are labeled at the frame level. In some previous works~\cite{Lu-ICCV-2013,Zhang-PR-2016}, the first $15$ minutes ($22500$ frames) in both videos are kept for training, although others~\cite{Cheng-CVPR-2015} have used more than half of the video for training.

\noindent
{\bf UCSD.}
The UCSD Pedestrian data set~\cite{Mahadevan-CVPR-2010} is perhaps one of the most challenging anomaly detection data sets. It includes two subsets, namely Ped1 and Ped2. Ped1 contains $34$ training and $36$ test videos with a frame resolution of $238 \times 158$ pixels. There are $6800$ frames for training and $7200$ for testing. Pixel-level anomaly labels are provided for only $10$ test videos in Ped1. All the $36$ test videos are annotated at the frame-level. Ped2 contains $16$ training and $12$ test videos, and the frame resolution is $360 \times 240$ pixels. There are $2550$ frames for training and $2010$ for testing. Although Ped2 contains pixel-level as well as frame-level annotations for all the test videos, most previous works~\cite{Cong-CVPR-2011,Lu-ICCV-2013,Ren-BMVC-2015,Xu-BMVC-2015,Zhang-PR-2016} have reported the pixel-level performance only for Ped1. The videos illustrate various crowded scenes, and anomalies are bicycles, vehicles, skateboarders and wheelchairs crossing pedestrian areas.

\noindent
{\bf UMN.}
The UMN Unusual Crowd Activity data set~\cite{Mehran-CVPR-2009} consists of three different crowded scenes, each with $1453$, $4144$, and $2144$ frames, respectively. The resolution of each frame is $320 \times 240$ pixels. In the normal settings people walk around in the scene, and the abnormal behavior is defined as people running in all directions.

\subsection{Evaluation}

We employ ROC curves and the corresponding \emph{area under the curve} (AUC) as the evaluation metric, computed with respect to ground truth frame-level annotations, and, when available (Avenue and UCSD), pixel-level annotations. We define the frame-level and pixel-level AUC as~\cite{Cong-CVPR-2011, Giorno-ECCV-2016,Lu-ICCV-2013,Mahadevan-CVPR-2010} and others. At the frame-level, a frame is considered a correct detection if it contains at least one abnormal pixel. At the pixel-level, the corresponding frame is considered as being correctly detected if more than $40\%$ of truly anomalous pixels are detected. We use the same approach as~\cite{Giorno-ECCV-2016,Lu-ICCV-2013} to compute the pixel-level AUC. The frame-level scores produced by our framework are assigned to the corresponding spatio-temporal cubes. The results are smoothed with the same filter used by~\cite{Giorno-ECCV-2016,Lu-ICCV-2013} in order to obtain our final pixel-level detections. Although many works~\cite{Cong-CVPR-2011,Dutta-AAAI-2015,Lu-ICCV-2013,Mahadevan-CVPR-2010,Xu-BMVC-2015,Zhang-PR-2016} include the Equal Error Rate (EER) as evaluation metric, we agree with~\cite{Giorno-ECCV-2016} that metrics such as the EER can be misleading in a realistic anomaly detection setting, in which abnormal events are expected to be very rare. Thus, we do not use the EER in our evaluation.

\subsection{Implementation Details}

We extract motion and appearance features from the test video sequences. We use the code available online at {https://alliedel.github.io/anomalydetection/} to compute the 3D motion gradients. For the appearance features, we consider the pre-trained VGG-f~\cite{Chatfield-BMVC-14} model provided in MatConvNet~\cite{matconvnet}. To detect changes, we employ the Logistic Regression implementation from VLFeat~\cite{vedaldi-vlfeat-2008}. In all the experiments, we set the regularization parameter of Logistic Regression to $0.1$, and we use the same window size as~\cite{Giorno-ECCV-2016}, namely $w=10$. We use the same parameters for both motion and appearance features.

In Table~\ref{tab_prelim_results}, we present preliminary results on five test videos from the Avenue data set to motivate our parameter and implementation choices. Regarding the CNN features, we show that slightly better results can be obtained with the \emph{conv5} features rather than the \emph{fc6} or \emph{fc7} features. An improvement of $2.5\%$ is obtained when we include unmasking. In the unmasking procedure, we use $k=10$ loops and eliminate the best $m=50$ features (top $25$ weighted as positive and top $25$ weighted as negative). A performance gain of $1.4\%$ can also be achieved when we divide the frames into $2\times 2$ bins instead of processing the entire frames. As for the stride, we present results with choices for $s \in \{1,2,5,10\}$. The time increases as we apply unmasking and spatial bins, but we can compensate by increasing the stride. We can observe that strides up to $10$ frames do not imply a considerable decrease in terms of frame-level or pixel-level AUC. Thus, we can set the stride to $5$ for an optimal trade-off between accuracy and speed. We show that very good results can also be obtained with motion features. In the end, we combine the two kinds of features and reach our best frame-level AUC ($82.6\%$). For the speed evaluation, we independently measure the time required to extract features and the time required to predict the anomaly scores on a computer with Intel Core i7 2.3 GHz processor and 8 GB of RAM using a single core. We present the number of frames per second (FPS) in Table~\ref{tab_prelim_results}. Using two cores, one for feature extraction and one for change detection by unmasking, our final model is able to process the videos at nearly $20$ FPS. For the rest of the experiments, we show results with both kinds of features using a stride of $5$ and bins of $2 \times 2$, and perform change detection by unmasking.

\subsection{Results on the Avenue Data Set}

\begin{table}[t]
\small{
\begin{center}
\begin{tabular}{|l|c|c|}
\hline
Method 																& Frame AUC    	& Pixel AUC 	\\
\hline
\hline
Lu et al.~\cite{Lu-ICCV-2013}								& $80.9\%$			& $92.9\%$ \\
Del Giorno et al.~\cite{Giorno-ECCV-2016}			& $78.3\%$			& $91.0\%$ \\
\hline
Ours (conv5)														& $80.5\%$			& $92.9\%$ \\
Ours (3D gradients)												& $80.1\%$			& $93.0\%$ \\
Ours (late fusion)													& $80.6\%$			& $93.0\%$ \\
\hline
\end{tabular}
\end{center}
}
\vspace*{-0.1cm}
\caption{Abnormal event detection results in terms of frame-level and pixel-level AUC on the Avenue data set. Our unmasking framework is compared with a state-of-the-art unsupervised approach~\cite{Giorno-ECCV-2016} as well as a supervised one~\cite{Lu-ICCV-2013}.}
\label{tab_Avenue}
\vspace*{-0.2cm}
\end{table}

\begin{figure}
\begin{center}
\includegraphics[width=1.0\columnwidth]{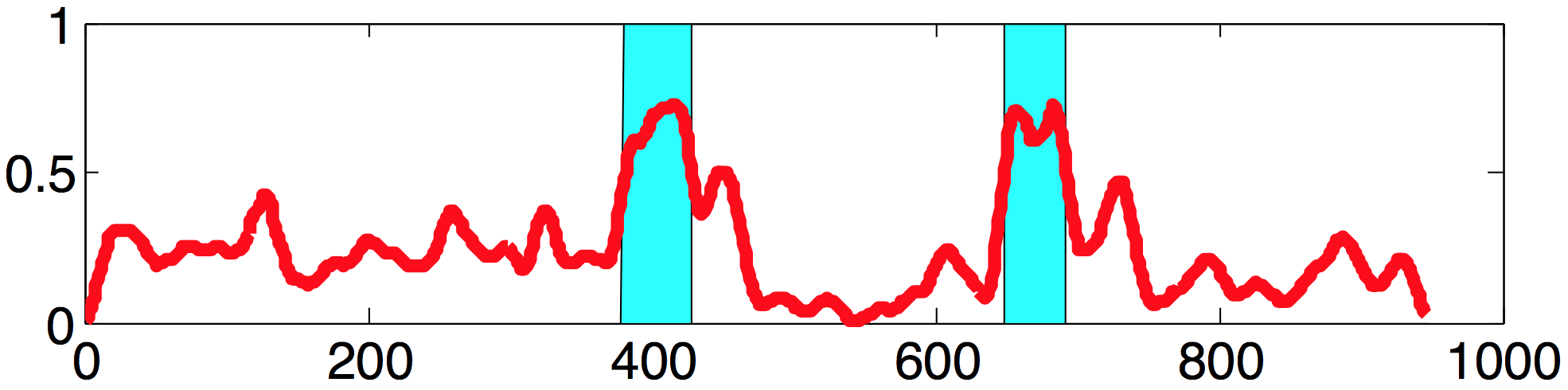}
\end{center}
\vspace*{-0.3cm}
\caption{Frame-level anomaly detection scores (between $0$ and $1$) provided by our unmasking framework based on the late fusion strategy, for test video 4 in the Avenue data set. The video has $947$ frames. Ground-truth abnormal events are represented in cyan, and our scores are illustrated in red. Best viewed in color.}
\label{fig_Avenue_vid4}
%\vspace*{-0.3cm}
\end{figure}

\begin{figure}
\begin{center}
\includegraphics[width=0.6\columnwidth]{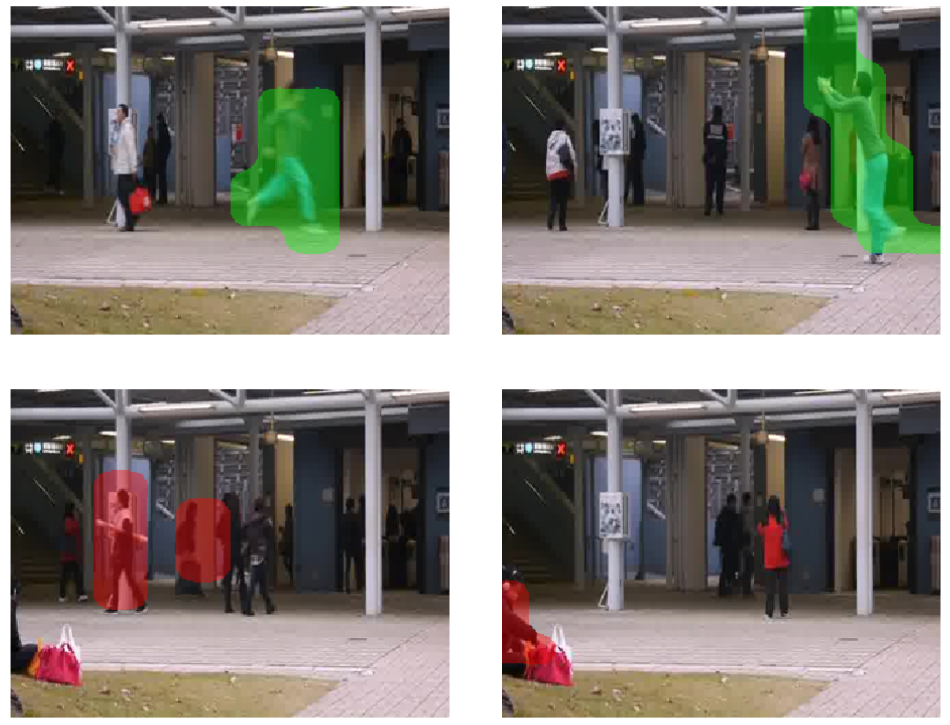}
\end{center}
\vspace*{-0.3cm}
\caption{True positive (top row) versus false positive (bottom row) detections of our unmasking framework based on the late fusion strategy. Examples are selected from the Avenue data set. Best viewed in color.}
\label{fig_Avenue_pos_neg}
\vspace*{-0.3cm}
\end{figure}

We first compare our unmasking framework based on several types of features with an unsupervised approach~\cite{Giorno-ECCV-2016} as well as a supervised one~\cite{Lu-ICCV-2013}. The frame-level and pixel-level AUC metrics computed on the Avenue data set are presented in Table~\ref{tab_Avenue}. Compared to the state-of-the-art unsupervised method~\cite{Giorno-ECCV-2016}, our framework brings an improvement of $2.3\%$, in terms of frame-level AUC, and an improvement of $2.0\%$, in terms of pixel-level AUC. The results are even more impressive, considering that our framework processes the video online, while the approach proposed in~\cite{Giorno-ECCV-2016} works only in offline mode. Moreover, our frame-level and pixel-level AUC scores reach about the same level as the supervised method~\cite{Lu-ICCV-2013}. Overall, our results on the Avenue data set are noteworthy.

Figure~\ref{fig_Avenue_vid4} illustrates the frame-level anomaly scores, for test video 4 in the Avenue data set, produced by our unmasking framework based on combining motion and appearance features using a late fusion strategy. According to the ground-truth anomaly labels, there are two abnormal events in this video. In Figure~\ref{fig_Avenue_vid4}, we notice that our scores correlate well to the ground-truth labels, and we can easily identify both abnormal events by setting a threshold of around $0.5$, without including any false positive detections. However, using this threshold there are some false positive detections on other test videos from the Avenue data set. We show some examples of true positive and false positive detections in Figure~\ref{fig_Avenue_pos_neg}. The true positive abnormal events are \emph{a person running} and \emph{a person throwing an object}, while false positive detections are \emph{a person holding a large object} and \emph{a person sitting on the ground}. 

\subsection{Results on the Subway Data Set}

\begin{table}[t]
\small{
\begin{center}
\begin{tabular}{|l|c|c|}
\hline
Method 																& \multicolumn{2}{|c|}{Frame AUC} \\
\cline{2-3}
			 																& Entrance gate  		& Exit gate\\
\hline
\hline
Cong et al.~\cite{Cong-CVPR-2011}					& $80.0\%$			& $83.0\%$ \\
Del Giorno et al.~\cite{Giorno-ECCV-2016}			& $69.1\%$			& $82.4\%$ \\
\hline
Ours (conv5)														& $69.5\%$			& $84.7\%$\\
Ours (3D gradients)												& $71.3\%$			& $86.3\%$ \\
Ours (late fusion)													& $70.6\%$			& $85.7\%$ \\
\hline
\end{tabular}
\end{center}
}
\vspace*{-0.1cm}
\caption{Abnormal event detection results in terms of frame-level AUC on the Subway data set. Our unmasking framework is compared with a state-of-the-art unsupervised approach~\cite{Giorno-ECCV-2016} as well as a supervised one~\cite{Cong-CVPR-2011}.}
\label{tab_Subway}
\vspace*{-0.1cm}
\end{table}

\begin{figure}
\begin{center}
\includegraphics[width=0.6\columnwidth]{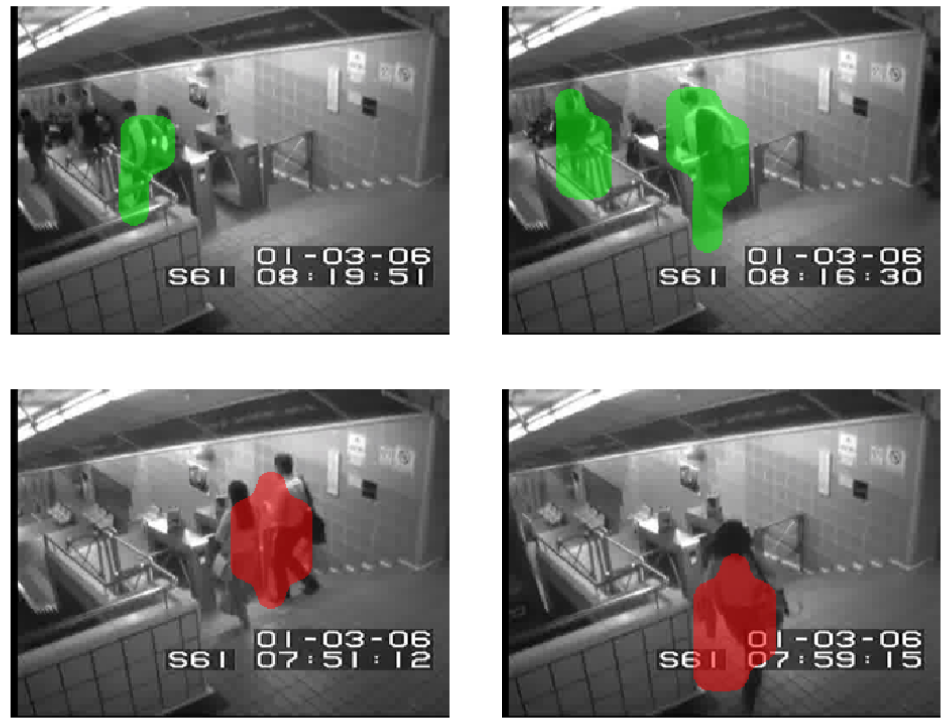}
\end{center}
\vspace*{-0.3cm}
\caption{True positive (top row) versus false positive (bottom row) detections of our unmasking framework based on the late fusion strategy. Examples are selected from the Subway Entrance gate. Best viewed in color.}
\label{fig_Subway_pos_neg}
\vspace*{-0.3cm}
\end{figure}

On the Subway data set, we compare our unmasking framework with two approaches, an unsupervised one~\cite{Giorno-ECCV-2016} and a supervised one~\cite{Cong-CVPR-2011}. The comparative results are presented in Table~\ref{tab_Subway}. On this data set, we generally obtain better results by using motion features rather than appearance features. Our late fusion strategy is not able to bring any improvements. Nevertheless, for each and every type of features, we obtain better results than the state-of-the-art unsupervised approach~\cite{Giorno-ECCV-2016}. When we combine the features, our improvements are $1.5\%$ on the Entrance gate video and $3.3\%$ on the Exit gate video. Remarkably, we even obtain better results than the supervised method~\cite{Cong-CVPR-2011} on the Exit gate video. On the other hand, our unsupervised approach, as well as the approach of Del Giorno et al.~\cite{Giorno-ECCV-2016}, obtains much lower results on the Entrance gate video.

Although there are many works that used the Subway data set in the experiments~\cite{Cheng-CVPR-2015,Cong-CVPR-2011,Dutta-AAAI-2015,Lu-ICCV-2013,Zhang-PR-2016}, most of these works~\cite{Dutta-AAAI-2015,Lu-ICCV-2013,Zhang-PR-2016} did not use the frame-level AUC as evaluation metric. Therefore, we excluded these works from our comparison presented in Table~\ref{tab_Subway}. However, there is a recent work~\cite{Cheng-CVPR-2015} that provides the frame-level AUC, but it uses only $47\%$ of the Entrance gate video for testing. For a fair comparison, we evaluated our unmasking framework based on the late fusion strategy in their setting, and obtained a frame-level AUC of $78.1\%$. Our score is nearly $14.6\%$ lower than the score of $92.7\%$ reported in~\cite{Cheng-CVPR-2015}, confirming that there is indeed a significant performance gap between supervised and unsupervised methods on the Entrance gate video. Nevertheless, in Figure~\ref{fig_Subway_pos_neg}, we can observe some interesting qualitative results obtained by our framework on the Entrance gate video. The true positive abnormal events are \emph{a person sidestepping the gate} and \emph{a person jumping over the gate}, while false positive detections are \emph{two persons walking synchronously} and \emph{a person running} to catch the train. 

\subsection{Results on the UCSD Data Set}

\begin{table}[t]
\small{
\begin{center}
\begin{tabular}{|l|c|c|c|}
\hline
Method 																& \multicolumn{2}{|c|}{Ped1} 			& Ped2\\
\cline{2-4}
			 																& Frame   			& Pixel  				& Frame \\
			 																& AUC   				& AUC  				& AUC \\
\hline
\hline
Kim et al.~\cite{Kim-CVPR-2009}						& $59.0\%$			& $20.5\%$			& $69.3\%$ \\
Mehran et al.~\cite{Mehran-CVPR-2009}				& $67.5\%$			& $19.7\%$				& $55.6\%$ \\
Mahadevan et al.~\cite{Mahadevan-CVPR-2010}	& $81.8\%$			& $44.1\%$	 		& $82.9\%$	 \\
Cong et al.~\cite{Cong-CVPR-2011}					& -						& $46.1\%$ 		& - \\
Saligrama et al.~\cite{Saligrama-CVPR-2012}		& $92.7\%$			& -	 					& - \\
Lu et al.~\cite{Lu-ICCV-2013}								& $91.8\%$			& $63.8\%$			& - \\
Ren et al.~\cite{Ren-BMVC-2015}						& $70.7\%$			& $56.2\%$			& - \\
Xu et al.~\cite{Xu-BMVC-2015}							& $92.1\%$			& $67.2\%$ 		& $90.8\%$	 \\
Zhang et al.~\cite{Zhang-PR-2016}						& $87.0\%$			& $77.0\%$ 		& $91.0\%$	 \\
Sun et al.~\cite{Sun-PR-2017}							& $93.8\%$			& $65.1\%$ 		& $94.1\%$	 \\
\hline
Ours (conv5)														& $68.4\%$			& $52.5\%$ 		& $82.1\%$	 \\
Ours (3D gradients)												& $67.8\%$			& $52.3\%$ 		& $81.3\%$	 \\
Ours (late fusion)													& $68.4\%$			& $52.4\%$ 		& $82.2\%$	 \\
\hline
\end{tabular}
\end{center}
}
\vspace*{-0.1cm}
\caption{Abnormal event detection results in terms of frame-level and pixel-level AUC on the UCSD data set. Our unmasking framework is compared with several state-of-the-art supervised methods~\cite{Cong-CVPR-2011,Kim-CVPR-2009,Lu-ICCV-2013,Mahadevan-CVPR-2010,Mehran-CVPR-2009,Ren-BMVC-2015,Saligrama-CVPR-2012,Sun-PR-2017,Xu-BMVC-2015,Zhang-PR-2016}.}
\label{tab_UCSD}
\vspace*{-0.5cm}
\end{table}

\begin{figure}
\vspace*{-0.05cm}
\begin{center}
\includegraphics[width=0.90\columnwidth]{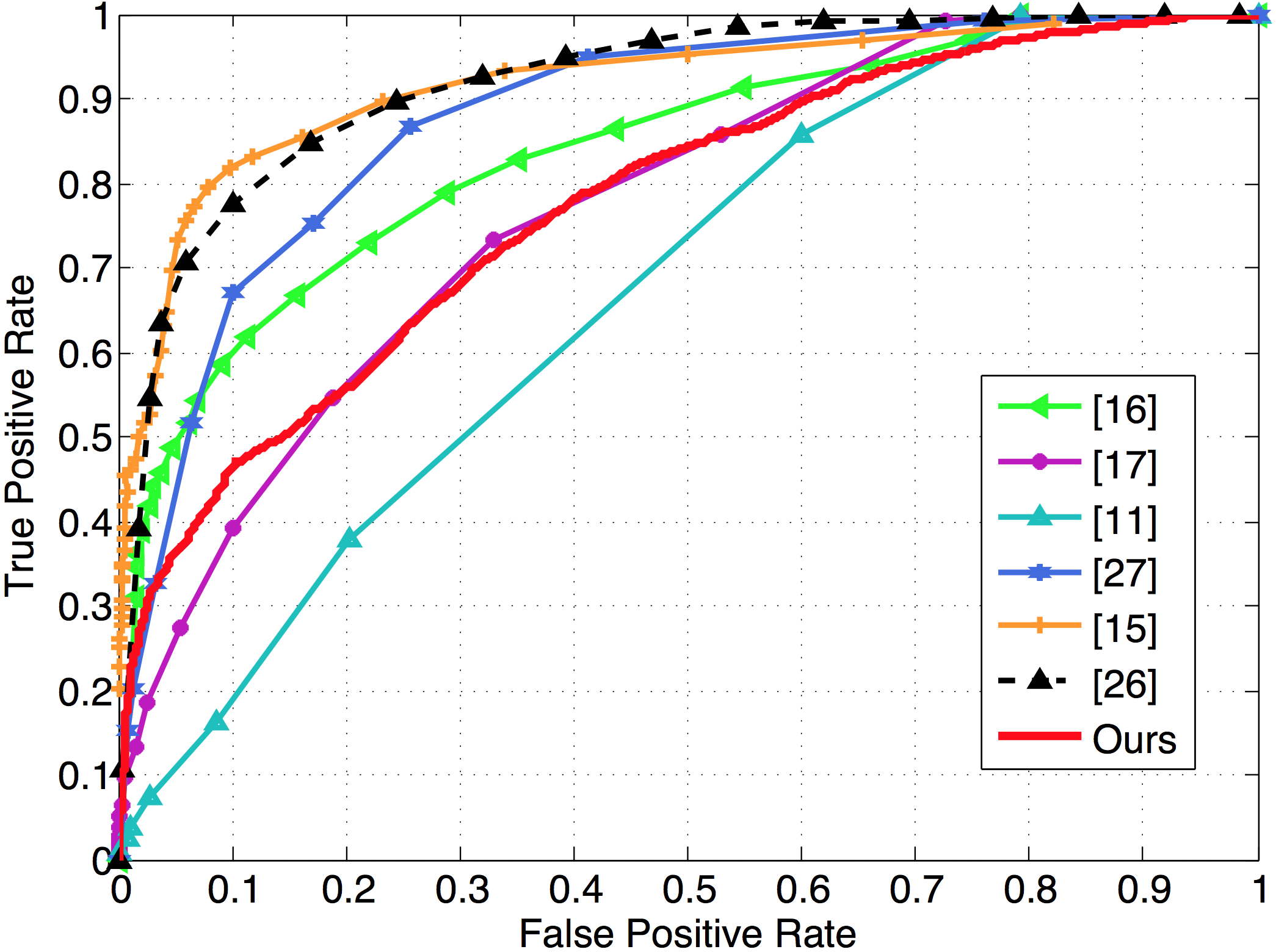}
\end{center}
\vspace*{-0.2cm}
\caption{Frame-level ROC curves of our framework versus~\cite{Kim-CVPR-2009,Lu-ICCV-2013,Mahadevan-CVPR-2010,Mehran-CVPR-2009,Xu-BMVC-2015,Zhang-PR-2016} on UCSD Ped1. Best viewed in color.}
\label{fig_UCSD_frame_ROC}
\vspace*{-0.1cm}
\end{figure}

\begin{figure}
\begin{center}
\includegraphics[width=0.90\columnwidth]{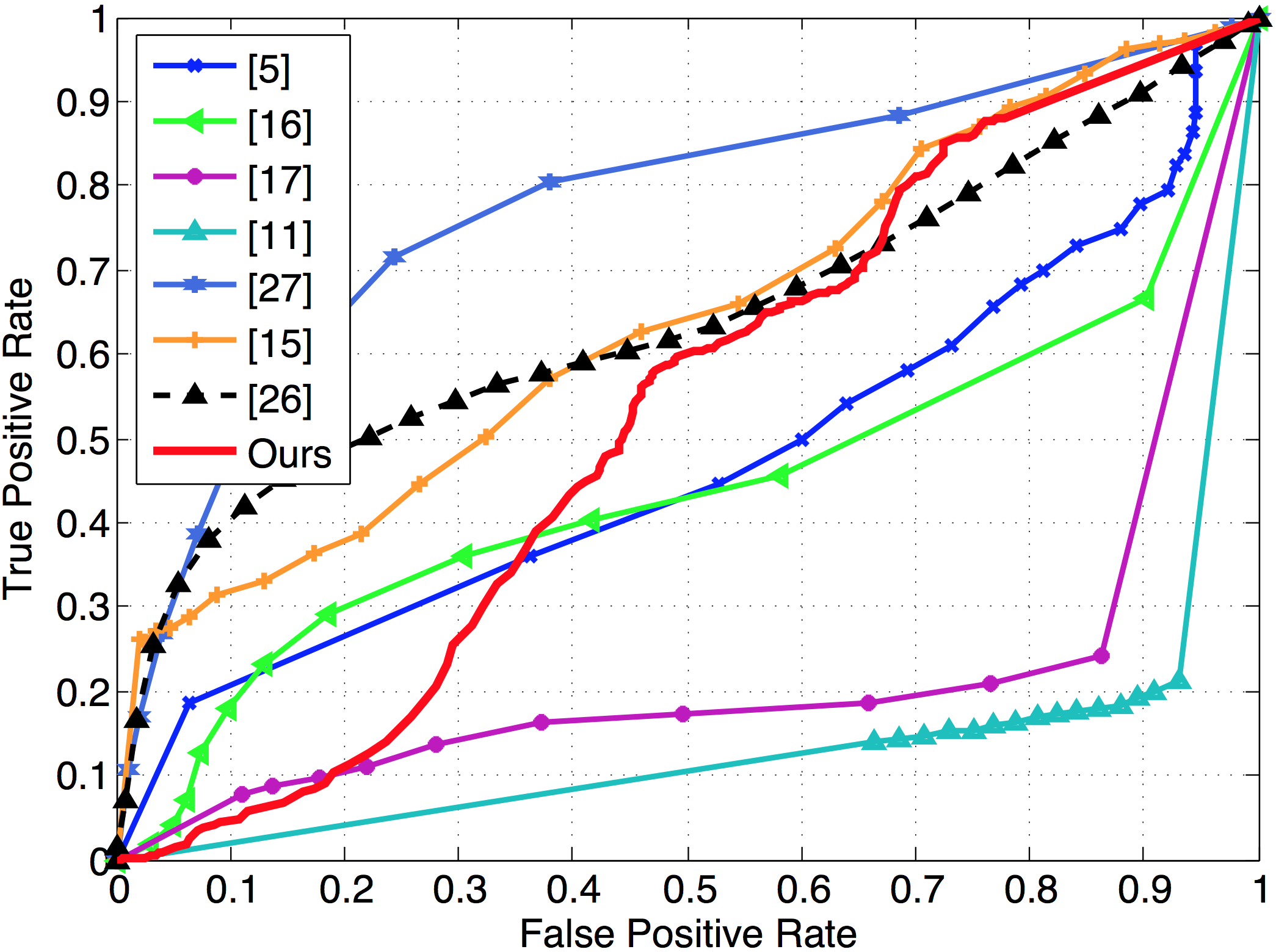}
\end{center}
\vspace*{-0.2cm}
\caption{Pixel-level ROC curves of our framework versus~\cite{Cong-CVPR-2011,Kim-CVPR-2009,Lu-ICCV-2013,Mahadevan-CVPR-2010,Mehran-CVPR-2009,Xu-BMVC-2015,Zhang-PR-2016} on UCSD Ped1. Best viewed in color.}
\label{fig_UCSD_pixel_ROC}
%\vspace*{-0.3cm}
\end{figure}

\begin{figure}
\begin{center}
\includegraphics[width=0.6\columnwidth]{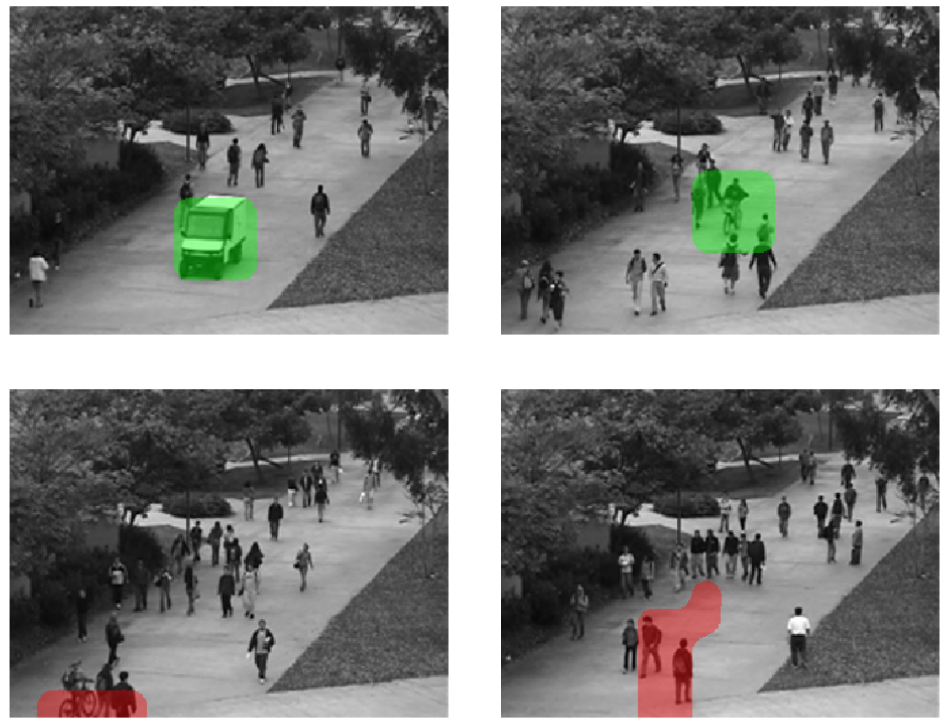}
\end{center}
\vspace*{-0.2cm}
\caption{True positive (top row) versus false positive (bottom row) detections of our unmasking framework based on the late fusion strategy. Examples are selected from the UCSD data set. Best viewed in color.}
\label{fig_UCSD_pos_neg}
\vspace*{-0.4cm}
\end{figure}

Del Giorno et al.~\cite{Giorno-ECCV-2016} have excluded the UCSD data set from their experiments because nearly half of the frames in each test video contain anomalies, and, they expect abnormal events to be rare in their setting. Although we believe that \emph{our approach would perform optimally in a similar setting}, we still compare our unsupervised approach with several state-of-the-art methods that require training data~\cite{Cong-CVPR-2011,Kim-CVPR-2009,Lu-ICCV-2013,Mahadevan-CVPR-2010,Mehran-CVPR-2009,Ren-BMVC-2015,Saligrama-CVPR-2012,Sun-PR-2017,Xu-BMVC-2015,Zhang-PR-2016}. In Table~\ref{tab_UCSD}, we present the frame-level and pixel-level AUC for Ped1, and the frame-level AUC for Ped2. In terms of frame-level AUC, we obtain better results than two supervised methods~\cite{Kim-CVPR-2009,Mehran-CVPR-2009}. In terms of pixel-level AUC, we obtain better results than four methods~\cite{Cong-CVPR-2011,Kim-CVPR-2009,Mahadevan-CVPR-2010,Mehran-CVPR-2009}. On Ped1, our results are only $3$ or $4\%$ lower than to those reported by Ren et al.~\cite{Ren-BMVC-2015}, while more recent supervised approaches achieve much better results~\cite{Sun-PR-2017,Zhang-PR-2016}. As most of the previous works, we have included the frame-level and pixel-level ROC curves for Ped1, to give the means for a thorough comparison with other approaches. Figure~\ref{fig_UCSD_frame_ROC} shows the frame-level ROC corresponding to the frame-level AUC of $68.4\%$ reached by our unmasking framework based on late fusion, while Figure~\ref{fig_UCSD_pixel_ROC} shows the pixel-level ROC corresponding to the pixel-level AUC of $52.4\%$ obtained with the same configuration for our approach.
Some qualitative results of our unsupervised framework based on late fusion are illustrated in Figure~\ref{fig_UCSD_pos_neg}. The true positive abnormal events are \emph{a car intruding a pedestrian area} and \emph{a bicycle rider intruding a pedestrian area}, while false positive detections are \emph{a bicycle rider and two persons walking synchronously} and \emph{two persons walking in opposite directions}. 

\vspace*{-0.1cm}
\subsection{Results on the UMN Data Set}

\begin{table}[t]
\small{
\begin{center}
\begin{tabular}{|l|c|c|c|c|}
\hline
Method 																& \multicolumn{4}{|c|}{Frame AUC} \\
\cline{2-5}
			 																& \multicolumn{3}{|c|}{Scene} 									& All \\
\cline{2-4}
			 																&  1   					&  2 						&  3 						& scenes \\
\hline
\hline
Mehran et al.~\cite{Mehran-CVPR-2009}				& -						& - 						& -						& $96.0\%$\\
Cong et al.~\cite{Cong-CVPR-2011}					& $99.5\%$			& $97.5\%$ 		& $96.4\%$			& $97.8\%$\\
Saligrama et al.~\cite{Saligrama-CVPR-2012}		& -						& -	 					& -						& $98.5\%$\\
Zhang et al.~\cite{Zhang-PR-2016}						& $99.2\%$			& $98.3\%$ 		& $98.7\%$			& $98.7\%$\\
Sun et al.~\cite{Sun-PR-2017}							& $99.8\%$			& $99.3\%$ 		& $99.9\%$			& $99.7\%$\\
Del Giorno et al.~\cite{Giorno-ECCV-2016}			& -						& -						& - 						& $91.0\%$\\
\hline
Ours (conv5)														& $98.9\%$			& $86.5\%$ 		& $98.5\%$			& $94.5\%$\\
Ours (3D gradients)												& $99.7\%$			& $84.9\%$ 		& $97.4\%$			& $94.0\%$\\
Ours (late fusion)													& $99.3\%$			& $87.7\%$ 		& $98.2\%$			& $95.1\%$\\
\hline
\end{tabular}
\end{center}
}
\vspace*{-0.1cm}
\caption{Abnormal event detection results in terms of frame-level AUC on the UMN data set. Our unmasking framework is compared with several state-of-the-art supervised methods~\cite{Cong-CVPR-2011,Mehran-CVPR-2009,Saligrama-CVPR-2012,Sun-PR-2017,Zhang-PR-2016} as well as an unsupervised approach~\cite{Giorno-ECCV-2016}.}
\label{tab_UMN}
\vspace*{-0.2cm}
\end{table}

\begin{figure}
\begin{center}
\includegraphics[width=1.0\columnwidth]{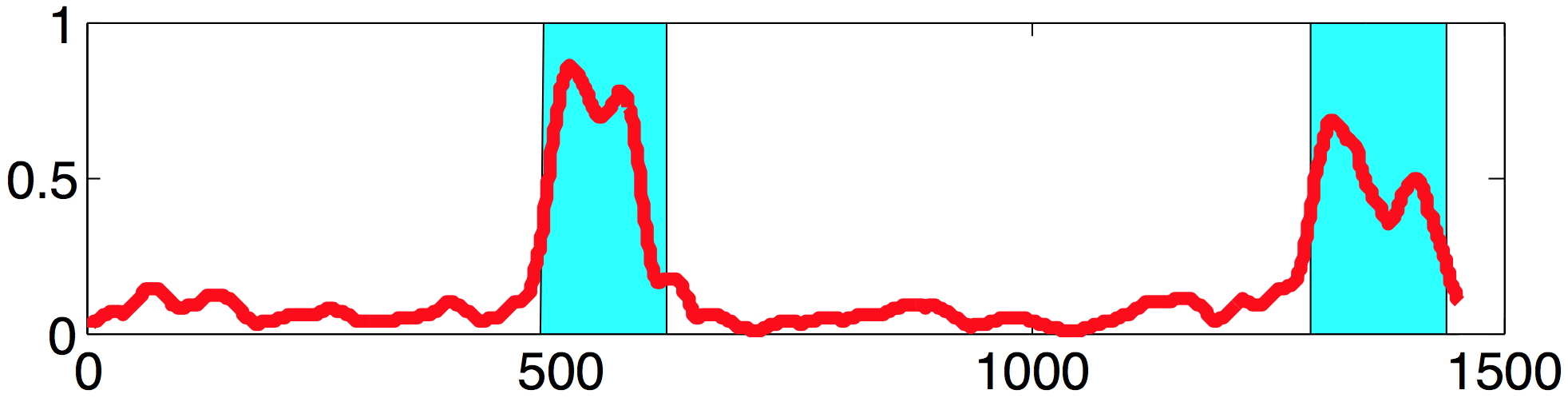}
\end{center}
\vspace*{-0.3cm}
\caption{Frame-level anomaly detection scores (between $0$ and $1$) provided by our unmasking framework based on the late fusion strategy, for the first scene in the UMN data set. The video has $1453$ frames. Ground-truth abnormal events are represented in cyan, and our scores are illustrated in red. Best viewed in color.}
\label{fig_UMN_scene1}
\vspace*{-0.1cm}
\end{figure}

\begin{figure}
\begin{center}
\includegraphics[width=0.6\columnwidth]{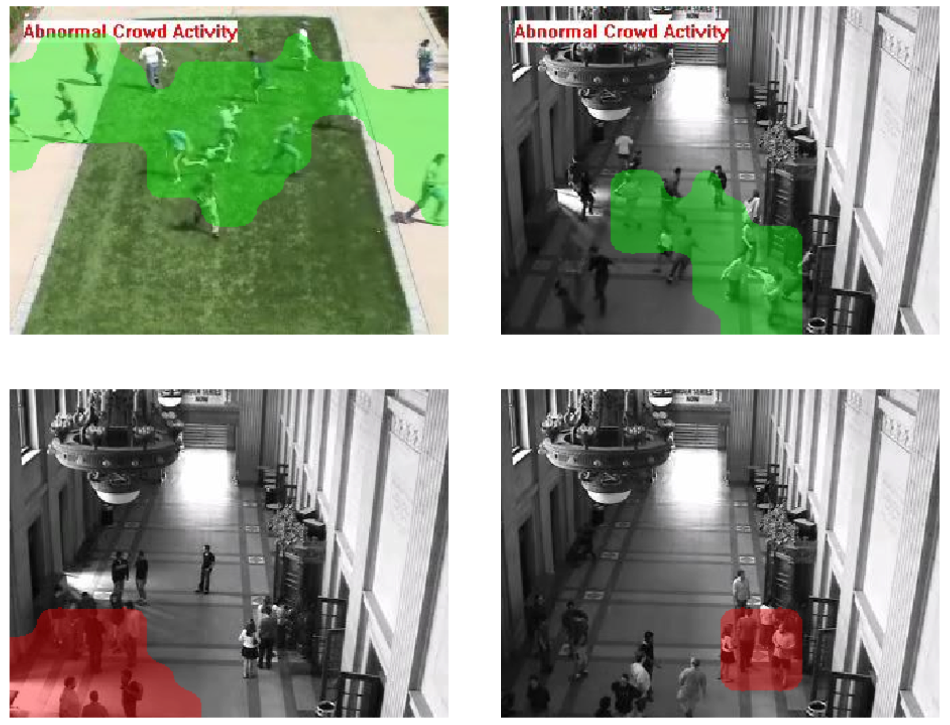}
\end{center}
\vspace*{-0.3cm}
\caption{True positive (top row) versus false positive (bottom row) detections of our unmasking framework based on the late fusion strategy. Examples are selected from the UMN data set. Best viewed in color.}
\label{fig_UMN_pos_neg}
\vspace*{-0.5cm}
\end{figure}

On the UMN data set, we compare our unmasking framework with a state-of-the-art unsupervised method~\cite{Giorno-ECCV-2016} and several supervised ones~\cite{Cong-CVPR-2011,Mehran-CVPR-2009,Saligrama-CVPR-2012,Sun-PR-2017,Zhang-PR-2016}. In Table~\ref{tab_UMN}, we present the frame-level AUC score for each individual scene, as well as the average score for all the three scenes. Compared to the unsupervised approach of Del Giorno et al.~\cite{Giorno-ECCV-2016}, we obtain an improvement of $4.1\%$. On the first scene, our performance is on par with the supervised approaches~\cite{Cong-CVPR-2011,Sun-PR-2017,Zhang-PR-2016}. As illustrated in Figure~\ref{fig_UMN_scene1}, our approach is able to correctly identify the two abnormal events in the first scene without any false positives, by applying a threshold of around $0.5$. On the last scene, the performance of our unmasking framework based on late fusion is less than $2\%$ lower than the best supervised approach~\cite{Sun-PR-2017}. Furthermore, we are able to surpass the performance reported in~\cite{Cong-CVPR-2011} for the third scene, by $1.8\%$. Our results are much worse on the second scene. We believe that the changes in illumination when people enter the room have a negative impact on our approach. The impact becomes more noticeable when we employ motion features alone, as the frame-level AUC is only $84.9\%$. Since the CNN features are more robust to illumination variations, we obtain a frame-level AUC of $86.5\%$. These observations are also applicable when we analyze the false positive detections presented in Figure~\ref{fig_UMN_pos_neg}. Indeed, the example in the bottom left corner of Figure~\ref{fig_UMN_pos_neg} illustrates that our method triggers a false detection when a significant amount of light enters the room as the door opens. The true positive examples in Figure~\ref{fig_UMN_pos_neg} represent \emph{people running around in all directions}.

\section{Conclusion and Future Work}
\label{sec_Conclusion}

In this work, we proposed a novel framework for abnormal event detection in video that requires no training sequences. Our framework is based on unmasking~\cite{Koppel-JMLR-2007}, a technique that has never been used in computer vision, as far as we know. We have conducted abnormal event detection experiments on four data sets in order to compare our approach with a state-of-the-art unsupervised approach~\cite{Giorno-ECCV-2016} and several supervised methods~\cite{Cong-CVPR-2011,Kim-CVPR-2009,Lu-ICCV-2013,Mahadevan-CVPR-2010,Mehran-CVPR-2009,Ren-BMVC-2015,Saligrama-CVPR-2012,Sun-PR-2017,Xu-BMVC-2015,Zhang-PR-2016}. The empirical results indicate that our approach gives better performance than the unsupervised method~\cite{Giorno-ECCV-2016} and some of the supervised ones~\cite{Cong-CVPR-2011,Kim-CVPR-2009,Mehran-CVPR-2009}. Unlike Del Giorno et al.~\cite{Giorno-ECCV-2016}, we can process the video online, without any accuracy degradation.

We have adopted a late fusion strategy to combine motion and appearance features, but we did not observe any considerable improvements when using this strategy. In future work, we aim at finding a better way of fusing motion and appearance features. Alternatively, we could develop an approach to train (unsupervised) deep features on a related task, e.g. action recognition, and use these features to represent both motion and appearance information.

\vspace*{0.3cm}
%\subsubsection*{Acknowledgments}
\noindent
{\bf Acknowledgments.}
% We thank reviewers for their helpful comments. 
Research supported by University of Bucharest, Faculty of Mathematics and Computer Science, through the 2017 Mobility Fund, and by SecurifAI through Project P/38/185 funded under %the Competitiveness Operational Programme 
POC-A1-A1.1.1-C-2015.

{\small
\bibliographystyle{ieee}
\bibliography{references}
}

\end{document}